%% file: main.tex
\def\year{2022}\relax
\documentclass[letterpaper]{article} % DO NOT CHANGE THIS
\usepackage{aaai22}  % DO NOT CHANGE THIS
\usepackage{times}  % DO NOT CHANGE THIS
\usepackage{helvet}  % DO NOT CHANGE THIS
\usepackage{courier}  % DO NOT CHANGE THIS
\usepackage[hyphens]{url}  % DO NOT CHANGE THIS
\usepackage{graphicx} % DO NOT CHANGE THIS
\usepackage{natbib}  % DO NOT CHANGE THIS AND DO NOT ADD ANY OPTIONS TO IT
\usepackage{caption} % DO NOT CHANGE THIS AND DO NOT ADD ANY OPTIONS TO IT
\DeclareCaptionStyle{ruled}{labelfont=normalfont,labelsep=colon,strut=off} % DO NOT CHANGE THIS
\usepackage{algorithm}
\usepackage{algorithmic}
\usepackage{multirow}
\usepackage{amsmath}
\usepackage{amssymb}
\usepackage{graphicx}
\usepackage{xcolor}
\usepackage{subfigure}
\usepackage{dsfont}
\usepackage{newfloat}
\usepackage{listings}
\floatstyle{ruled}
\newfloat{listing}{tb}{lst}{}
\floatname{listing}{Listing}
\title{Laneformer: Object-aware Row-Column Transformers for Lane Detection}
\author{Jianhua Han\textsuperscript{\rm 1}, Xiajun Deng\textsuperscript{\rm 2}, Xinyue Cai\textsuperscript{\rm 1}, Zhen Yang\textsuperscript{\rm 1}, Hang Xu\textsuperscript{\rm 1}, Chunjing Xu\textsuperscript{\rm 1}, Xiaodan Liang\textsuperscript{\rm 2}\thanks{Corresponding authors: liangxd9@mail.sysu.edu.cn}}
\affiliations{
    \textsuperscript{\rm 1}Huawei Noah's Ark Lab, \\ \textsuperscript{\rm 2}Shenzhen Campus of Sun Yat-sen University
}

\usepackage{bibentry}
\newlength\savewidth\newcommand\shline{\noalign{\global\savewidth\arrayrulewidth
  \global\arrayrulewidth 1pt}\hline\noalign{\global\arrayrulewidth\savewidth}}
\newcommand{\tablestyle}[2]{\setlength{\tabcolsep}{#1}\renewcommand{\arraystretch}{#2}\centering\footnotesize}

\begin{document}
\maketitle

%------------------------------------------------------------------------- 

\input{file/0-abstract}
\input{file/1-introduction}

\input{file/2-related}

\input{file/3-methods}
\input{file/4-experiment}
\input{file/5-conclusion}

\bibliography{main}

\input{file/6-appendix}

\end{document}

%% file: file/0-abstract.tex
% LZ version
\begin{abstract}
We present \textbf{Laneformer}, a conceptually simple yet powerful transformer-based architecture tailored for lane detection that is a long-standing research topic for visual perception in autonomous driving. The dominant paradigms rely on purely CNN-based architectures which often fail in incorporating relations of long-range lane points and global contexts induced by surrounding objects (e.g., pedestrians, vehicles). Inspired by recent advances of the transformer encoder-decoder architecture in various vision tasks, we move forwards to design a new end-to-end Laneformer architecture that revolutionizes the conventional transformers into better capturing the shape and semantic characteristics of lanes, with minimal overhead in latency. First, coupling with deformable pixel-wise self-attention in the encoder, Laneformer presents two new row and column self-attention operations to efficiently mine point context along with the lane shapes. Second, motivated by the appearing objects would affect the decision of predicting lane segments, Laneformer further includes the detected object instances as extra inputs of multi-head attention blocks in the encoder and decoder to facilitate the lane point detection by sensing semantic contexts. Specifically, the bounding box locations of objects are added into \textit{Key} module to provide interaction with each pixel and query while the ROI-aligned features are inserted into \textit{Value} module. Extensive experiments demonstrate our Laneformer achieves state-of-the-art performances on CULane benchmark, in terms of 77.1\% F1 score. We hope our simple and effective Laneformer will serve as a strong baseline for future research in self-attention models for lane detection.
\end{abstract}

%% file: file/1-introduction.tex
\section{Introduction}

\begin{figure}[htb]
		\begin{center}
            \includegraphics[width=0.9\linewidth]{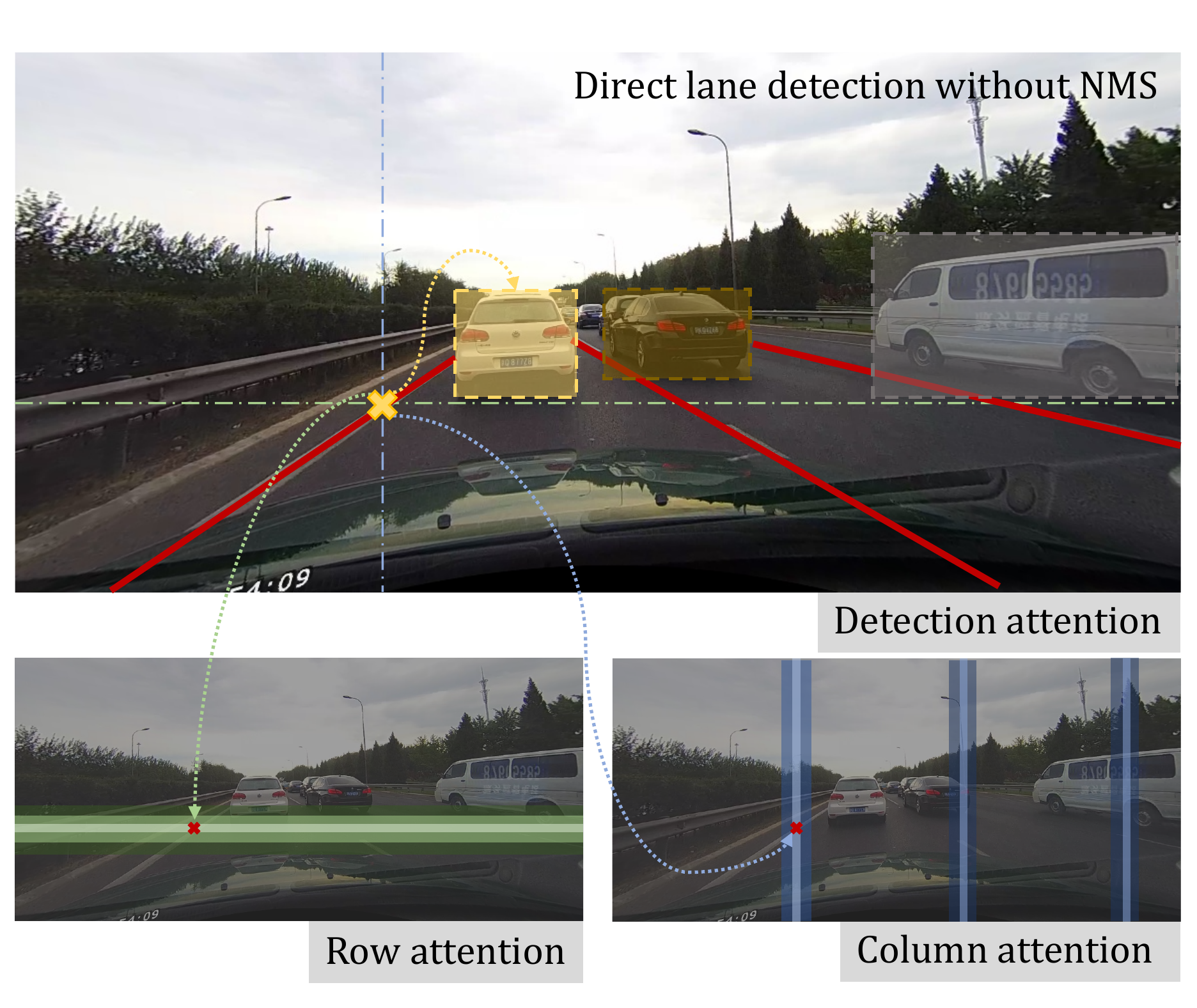}
		\end{center}
		\caption{Sketch map of proposed detection attention and row-column attention in Laneformer. Given the detected person and vehicle instances, detection attention is performed to capture the implicit relationship between them and lanes, e.g., lanes are more likely to appear next to cars. Row attention is proposed to catch the information from the nearby rows since pixels in the same lane will not be far from each other between adjacent rows. On the other hand, noticed that different columns may across different lanes, the knowledge sharing of these columns may capture different lane features to construct better representations.}
		\label{fig:long}
	\end{figure}
	
	Lane detection has been a long-standing task for visual perception in autonomous driving scenarios, targeting at precisely distinguishing lane segments from complicated road scenes\cite{hou2019learning, wang2000lane, narote2018review}.
	It plays a  crucial role towards a safe and reliable auto-driving system widely used in intelligent cars to assist drivers with modern technologies such as adaptive cruise control, lane departure warning, and traffic understanding. 
	
	The most state-of-the-art lane detection methods \cite{lee2021robust, xu2020curvelane, chen2019pointlanenet} take advantage of CNN architectures and exceed the traditional methods\cite{niu2016robust, narote2018review, wang2000lane} by a large margin.
    However, the existing CNN-based methods usually need complicated post-processing procedures like non-maximal suppression or clustering.
    In addition, the fixed receptive field of CNN architecture limits the ability to incorporate relations for long-range lane points, making them hard to capture well the characteristics of lanes since the shapes are conceptually long and thin.
    Several attention-based lane detection models~\cite{lee2021robust, tabelini2020keep, liu2021end} have been also proposed to capture the long-range information. 
    Nevertheless, the fixed attention routines can not adaptively fit the shape characteristic of lanes.
    Besides, complicated road scenes including different light, weather conditions and occlusions of surrounding objects further require the model with a stronger perception of global contexts.
    In addition to the point-wise context information, it is reasonable to assume that objects on the road (e.g., pedestrians, vehicles) have some implicit relations with surrounding lanes.
    As the key sub-module (e.g., person-vehicle detection) of autonomous driving systems usually co-exist with lane detection, the sub-module outputs may promote the performance of lane detection and improves the system safety. 
    However, none of the existing methods has considered incorporating semantics induced by detection results from a single and unified network view. 

    On the other hand, 	Transformer\cite{Vaswani2017}, a kind of encoder-decoder architecture, has shown surprisingly promising ability in dealing with tasks that require to capture global relations in nature language processing\cite{devlin2018bert, radford2018improving, young2018recent} and vision tasks such as image classification\cite{dosovitskiy2020image}, object detection\cite{carion2020end, zhu2020deformable} and image segmentation\cite{zheng2020rethinking, wang2020max}. 
    Particularly, \cite{liu2021end} proposed an LSTR model to predict lanes as polynomial functions using a transformer. Despite its benefit in providing rich global contexts in predicting lanes, the definition that regards lanes as polynomial functions has many drawbacks: 1) LSTR needs to formulate camera intrinsic and extrinsic parameters, which hinders the model from transferring to other datasets or train with combined datasets; 2) LSTR still lacks of explicitly modeling global semantic contexts into facilitating lane detection.
    
    To tackle the above-mentioned issues, we move forwards to design a new Laneformer architecture to better capture the shape characteristics and global semantic contexts of lanes using transformers (shown in Figure \ref{fig:long}). 
    Our Laneformer defines lanes as a series of points.
    Then, the lane-specific row-column attentions are proposed to efficiently mine point context along with the lane shapes.
    Concretely, we take each feature row as a token and perform row-to-row self-attention and do the same for each feature column to perform column-to-column self-attention.
    Moreover, motivated by the appearing objects that would affect lanes' predictions, Laneformer further includes the detected object instances as auxiliary inputs of multi-head attention blocks in encoder and decoder to facilitate the lane point detection by sensing semantic contexts. 
    To be specific, the bounding box locations of objects are added into the Key module to provide interaction with each pixel and query. At the same time, the ROI-aligned features are inserted into the Value module to provide detection information. 
    Considering that bounding box locations and ROI-aligned features contain limited information about object instances, we use confident scores and predicted categories of the detected outputs to further improve the model's performance.
    Bipartite matching is adopted to ensure one-to-one alignment between predictions and ground truths, which makes the Laneformer architecture eliminate additional post-processing procedures.
    
    Extensive experiments conducted on CULane\cite{pan2018spatial} and TuSimple\cite{TuSimple} benchmarks show that our Laneformer as an early transformer attempt on lane detection, achieves state-of-the-art performance on CULane with 77.1\% F1 score and superior 96.8\% accuracy on Tusimple, with about 50 FPS(frames-per-second) inference speed on V100. Besides, visualization of the learned attention map in transformer demonstrates that our Laneformer can incorporate relations of long-range lane points and global contexts induced by surrounding objects.
	
	To summarize, the main contributions can be listed: 
	
	\begin{itemize} 
        \item We design a new Laneformer architecture specified for lane detection, which accommodates the conventional transformers into capturing well the shape characteristics and semantic contexts of lanes.
		\item The lane-specific row-column self-attentions is proposed to mine point context along with the lane shapes with the prior that lanes are long and thin.
		\item The object instances by detection that usually co-exist in autonomous driving system are used as auxiliary inputs to sense object-wise semantic contexts.
		\item Experiments show that Laneformer achieves new state-of-the-art performance, in terms of 77.1\% F1 score on CULane and superior 96.8\% accuracy on Tusimple.
	\end{itemize}

%% file: file/2-related.tex
 \section{Related work}
\begin{figure*}[t!]
    \begin{center}
        \includegraphics[width=0.9\linewidth]{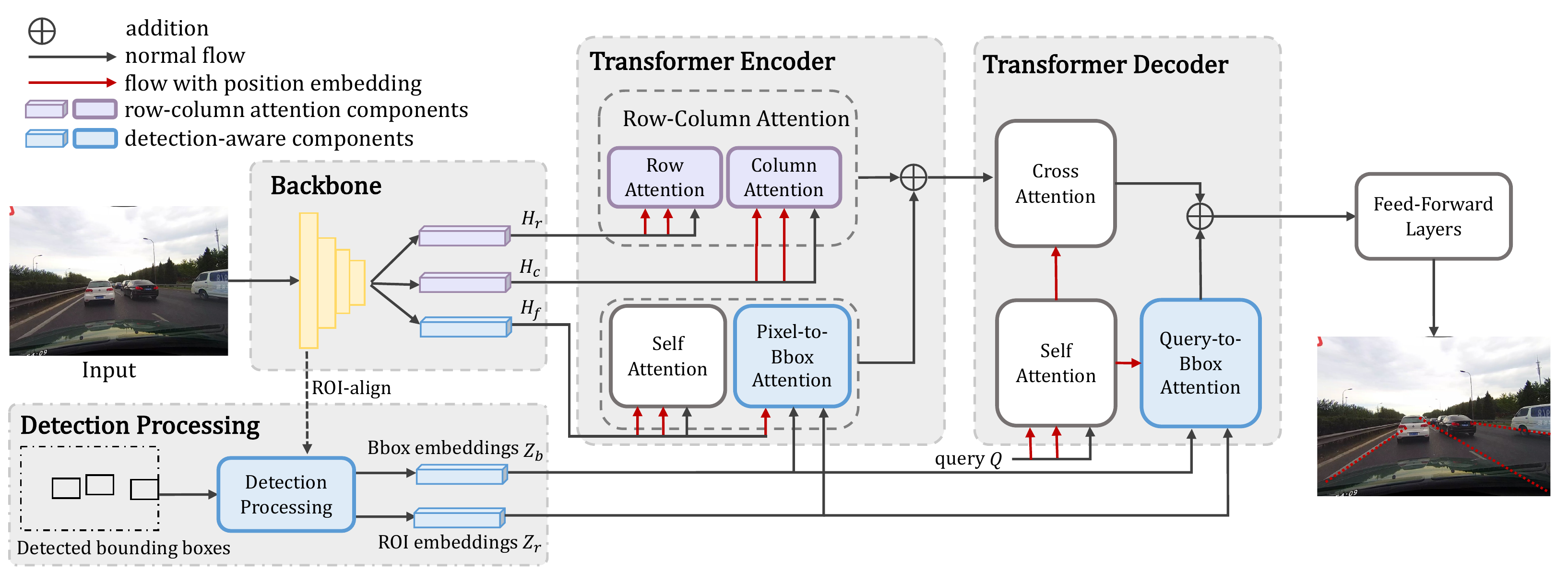}
    \end{center}
\caption{Overall Architecture. In Laneformer, backbone extracts backbone features $\mathbf{H_f}$, row features $\mathbf{H_r}$ and column features $\mathbf{H_c}$. Detection Processing module generates bounding box embeddings $\mathbf{Z_b}$ and ROI embeddings $\mathbf{Z_r}$ with detected bounding boxes and $\mathbf{H_f}$. In the encoder, row attention and column attention are performed on $\mathbf{H_r}$ and $\mathbf{H_c}$, added up with the pixel-to-Bbox attention performed on $\mathbf{H_f}$, $\mathbf{Z_b}$ and $\mathbf{Z_r}$ as memory input for decoder. In the decoder, traditional cross-attention and query-to-Bbox attention are performed and outputs are added up for the feed-forward layers to predict lanes.}
\label{fig:OverallArchitecture}
\end{figure*}
\noindent\textbf{Attention-free methods.} 
Before the advent of deep learning, traditional lane detection methods are usually based on hand-crafted low-level features \cite{chiu2005lane,lee2009effective,gonzalez2000lane}.
CNN architecture has then been adopted to extract advanced features in an end-to-end manner. 
Most lane detection methods follow pixel-level segmentation-based
approach \cite{DBLP:journals/corr/abs-1907-01294,DBLP:journals/corr/abs-1905-03704,DBLP:journals/corr/abs-1909-00798,DBLP:journals/corr/abs-1903-02193}. 
These approaches typically generate segmentation results with an encoder-decoder structure and then post-processing them via curve fitting and clustering. 
However, pixel-wise lane prediction methods usually require more computation and are also limited to the pre-defined number of lanes.
On the other hand, several works \cite{chen2019pointlanenet, li2019line, xu2020curvelane} follow traditional proposal-based diagrams by generating multiple point anchors and then predicting the relative distance between each lane point and the anchor point.
However, these existing CNN-based methods usually need complicated post-processing procedures like non-maximal suppression or clustering.
Besides, the fixed receptive field of CNN architecture limits the ability to incorporate relations for long-range lane points.
Therefore, our Laneformer utilizes a transformer to capture context information and bipartite matching to eliminate additional post-processing procedures.

\noindent\textbf{Attention-based methods.} Several attention-based lane detection models\cite{lee2021robust, tabelini2020keep, liu2021end} have been proposed to capture the long-range information. \cite{lee2021robust} propose a self-attention mechanism to predict the lanes' confidence along with the vertical and horizontal directions in an image. \cite{tabelini2020keep} proposes a novel anchor-based attention mechanism that aggregates global information named LaneATT. 
However, the fixed attention routines can not adaptively fit the shape characteristic of lanes.
Based on PolyLaneNet\cite{tabelini2020polylanenet}, LSTR\cite{liu2021end} is proposed to output polynomial parameters of a lane shape function by using a network built with a transformer to learn richer structures and context. 
Unlike LSTR, which assumes all lanes are parallel on the road and formulates it as a polynomial shape prediction problem, our  Laneformer directly outputs each lane's points to adapt to more complex lane detection scenarios. 
Besides, detected object instances are further included as auxiliary inputs of multi-head attention blocks in Laneformer to facilitate the lane point detection by sensing semantic contexts.

%% file: file/3-methods.tex
\section{Laneformer}

% In this section, we first introduce the structured representation of the lane.
% Then we give a detailed description of our Laneformer architecture and explain the construction of each module including detection processing module, row-column attention module and detection attention module.

\subsection{Lane Representation}
\label{sec:Lane Representation}

Similar to \cite{chen2019pointlanenet}, we define lanes as a series of 2D-points that can adapt to all kinds of lanes. 
Specially, we formulate a lane as $l = (\mathbf{X}, s, e)$, where $\mathbf{X}$ stands for corresponding x-coordinate set for 72 equally-spaced y-coordinates and $s$, $e$ denote for the start y-coordinate and end y-coordinate of the lane. 

\subsection{Architecture}
The overall architecture of Laneformer is demonstrated in Figure \ref{fig:OverallArchitecture}. 
It mainly consists of four modules: a CNN backbone to extract basic features, a detection processing module to handle outputs from person-vehicle detection module, a specially designed encoder and a decoder for lane detection.

Given an RGB image as input, our model first extracts backbone features with a ResNet \cite{he2016deep} backbone. We further get row features and column features by collapsing the column dimension and row dimension of backbone features, respectively.
At the same time, detected bounding boxes and their predicted scores and categories from the input image are acquired through a trained person-vehicle detection module. 
We use the bounding box locations to crop ROI-aligned features from backbone features mentioned above, followed by a 1-layer perceptron with ReLU activation to transform the ROI-aligned features to one-dimensional embeddings. 
Similarly, another 1-layer perceptron is applied on the detected bounding boxes to get one-dimensional bounding box embeddings.
In the encoder, row attention and column attention are performed on row features and column features respectively.
Meanwhile, backbone features will perform self-attention and pixel-to-Bbox attention with bounding box embeddings and ROI embeddings.
The outputs of row-column attention and pixel-to-Bbox attention are added up as the memory input for the following decoder.
In the decoder, learnable queries first perform self-attention to get query features. 
Then the query features together with the input memory will perform cross attention. Meanwhile, with the bounding box embeddings and ROI embeddings, query-to-Bbox attention can be applied.
Finally, outputs of the cross attention and query-to-Bbox attention are added up, and several feed-forward layers are utilized to predict lane points.
\subsection{Detection Processing}
After the acquisition of detected bounding boxes, predicted scores and predicted categories from a trained detector based on common Faster-RCNN\cite{ren2015faster} architecture, we propose a simple detection processing module to process them in order to better utilize the detection information.
First, we crop ROI-aligned features from backbone features $\mathbf{H_f} \in R^{h \times w \times d}$ for the bounding boxes of the detected objects, where $h, w$ and $d$ are the spatial sizes and the corresponding dimension of backbone features. 
Since the features we used are down-sampled, we need to rescale the bounding box locations by a specific ratio to crop out correct feature areas.
Considering that objects with higher confidence scores may supply more robust information while objects with lower scores may be noisy, we also use predicted scores as weight coefficients to multiply the ROI-aligned features and get weighted ROI-aligned features for each object.
After that, we pass through weighted ROI-aligned features into a 1-layer perceptron with output channel $d'$ followed by ReLu activation to get final ROI embeddings $\mathbf{Z_r}\in R^{M \times d'} $.
$M$ denotes the number of used detected bounding boxes.
If the number of detected bounding boxes is less than $M$, bounding boxes with random locations, categories and zero scores will be padded.
On the other hand, bounding boxes with lower scores will be excluded if the number of  detected bounding boxes is more than $M$.

In the meantime, with the prior that category information can help distinguish different objects, we further concatenate the predicted categories after bounding boxes.
Specifically, we use a four-dimensional vector to represent a bounding box and a one-hot vector with length 7 (1 for padded box and 6 for categories) to represent the corresponding category, which will result in an 11-dimensional vector after concatenation.
Similar to ROI features, a one-layer perceptron with ReLu activation is applied to get bounding box embeddings $\mathbf{Z_b} \in R^{M \times d'} $.
Finally, $\mathbf{Z_r}$ and $\mathbf{Z_b}$ are sent to preform Pixel-to-Bbox attention in encoder and Query-to-Bbox attention in decoder.

\begin{figure}[tb]
    \label{fig:attention map}
	\centering
	    \subfigure[\small{}Row-column attention.]{           \includegraphics[scale=0.4]{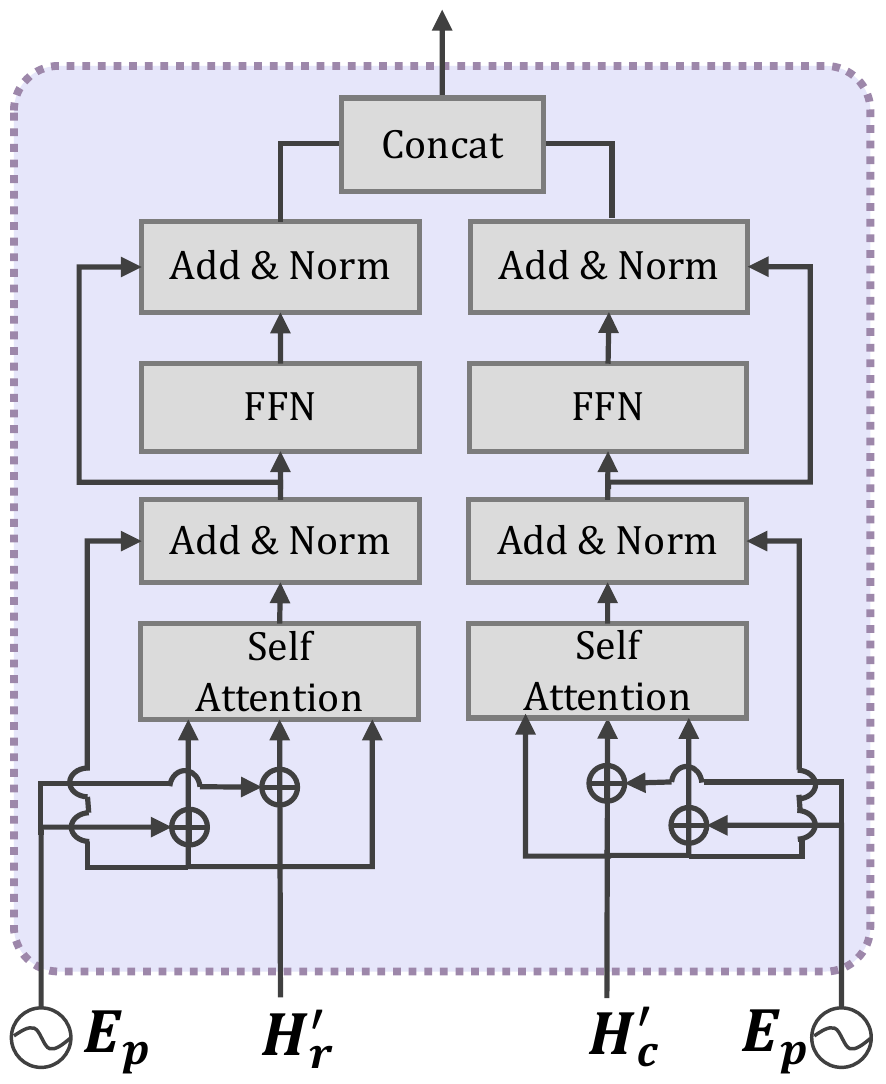} 
		\label{fig:row-column_attention}}
		\subfigure[\small{}Detection attention.]{                 \includegraphics[scale=0.4]{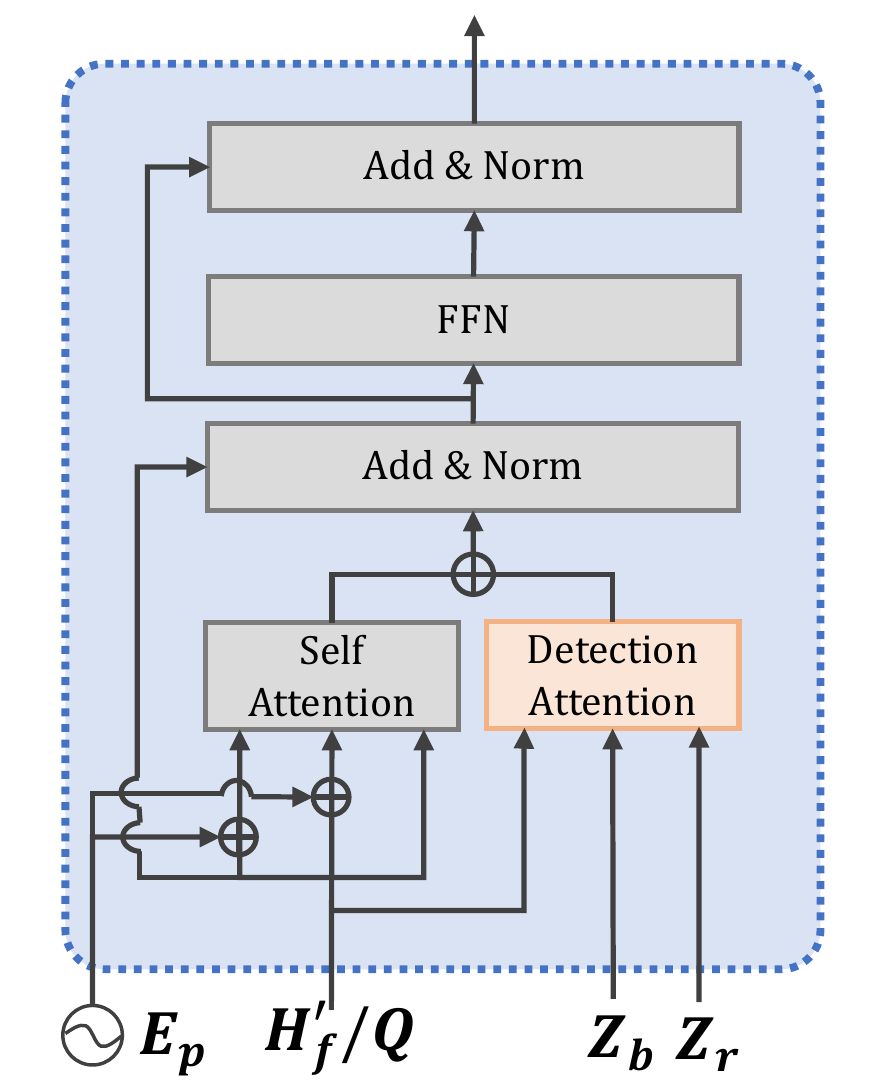} 
		\label{fig:detection-attention}}
	\caption{Detailed operations of the row-column attention and detection attention. $\bigoplus$ stands for the addition and $\mathbf{H_f'}/\mathbf{Q}$ denotes that the input can be either backbone feature $\mathbf{H_f'}$ in the encoder or query feature $\mathbf{Q}$ in the decoder.}%\xd{add details about the difference with previous attention}
	
\end{figure}
\subsection{Row-Column Attention}
\label{sec:Row-Column Encoder}
In the encoder of Laneformer, besides the traditional attention, we further propose the new row-column attention to efficiently mine point context along with the lane shapes.
The row-column attention consists of row attention that captures the relations between rows and column attention that mine the relations between columns.
As we all know, the traffic lane is a kind of object with unique shape characteristics. 
From the vertical view, lanes are long and thin, which means pixels in the same lane will not be far from each other between adjacent rows.
From the horizontal view, since different columns may across different lanes, the knowledge sharing of these columns may capture different lane features to construct better representations.

Given row features $\mathbf{H_r} \in R^{h \times 1 \times wd}$, where $h, w$ and $d$ are the spatial sizes and the the corresponding dimension of backbone features, we pass it through a linear transformation to reduce the last dimension to $d'$, which we denoted as $\mathbf{H_r'} \in R^{h \times 1 \times d'}$. 
Sinusoidal embeddings $\mathbf{E_p}$ is calculated according to the absolute positions to supply location information. 
Self-attention operations are applied with the inputs of $\mathbf{H_r'}$ and $\mathbf{E_p}$. Similarly, with the dimension-reduced column features $\mathbf{H_c'} \in R^{1 \times w \times d'}$ and its position embeddings, self-attention are performed between columns.
Finally, the outputs of row and column attention are added up as the memory input for decoder.
Details are shown in Figure \ref{fig:row-column_attention}.

\subsection{Detection Attention}
\label{sec:Detection Attention Module}

Apart from row-column attention, in both the encoder and decoder of Laneformer, we propose detection attention to mine the valuable information from detected surrounding objects.
It is straight-forward that lanes and objects on the road (e.g., pedestrians, vehicles) have some implicit relations with each other. 
For example, lanes are more likely to appear next to vehicles, while pedestrians are not supposed to walk between lanes. 
Besides, object detection module such as person-vehicle detection is always co-existed with lane detection in an auto-driving system, thus detected results can be acquired easily.
The detection attention module consists of two parts: 1) a Pixel-to-Bbox attention module in encoder to excavate the relevance between each pixel token of feature map and each detected objects; 2) a Query-to-Bbox attention module in decoder to find out which object should be paid more attention in order to help predict corresponding lanes. Details are illustrated in Figure \ref{fig:detection-attention}.

\subsubsection{Pixel-to-Bbox Attention}
\label{sec:Pixel-to-Bbox Attention}

% why here no indent?
In Laneformer encoder, we propose pixel-to-Bbox attention. Pixel-to-Bbox attention is designed for digging out relations between feature pixels and detected objects. In pixel-to-Bbox attention, each pixel in dimension-reduced backbone features $\mathbf{H_f'}$ is considered as a query token. Detection information including bounding box embeddings $\mathbf{Z_b}$ and ROI embeddings $\mathbf{Z_r}$ make up the Key module and Value module, respectively. The pixel-to-Bbox attention is defined as follows:

\begin{equation} \label{p2b-attention}
\mathbf{O_{p2b}} = softmax(\frac{\mathbf{H_{f}'}\mathbf{Z_{b}}^{T}}{\sqrt{d'}}) \cdot \mathbf{Z_{r}}
\end{equation}

Pixel-to-Bbox attention forces the model to learn which detected object a pixel should pay attention to so that the model can catch helpful context features.
The output of pixel-to-Bbox attention $\mathbf{O_{p2b}}$ are added up with the above row-column attention output and act as the memory input for the decoder.

\subsubsection{Query-to-Bbox Attention}
\label{sec:Query-to-Bbox Attention}
In Laneformer decoder, we propose query-to-Bbox attention. Query-to-Bbox attention on the other hand designed to mine relations between queries and detected objects. Query-to-Bbox attention is similar to Pixel-to-Bbox attention, while in here, queries are the learned embeddings $\mathbf{Q}$ with the size of $N \times d'$, where $N$ is the number of learned embeddings. Bounding box embeddings $\mathbf{Z_b}$ and ROI embeddings $\mathbf{Z_r}$ still act as Key module and Value module here. Similarly, the query-to-Bbox attention is defined as follows:

\begin{equation}
\mathbf{O_{q2b}} = softmax(\frac{\mathbf{Q}\mathbf{Z_{b}^T}}{\sqrt{d'}}) \cdot \mathbf{Z_{r}}
\end{equation}
This attention enables each query to focus on the instances near the lane it needs to predict. 
The output of query-to-Bbox attention $\mathbf{O_{q2b}}$ will be added up with the traditional cross-attention output, followed by several feed-forward features to get predicted lanes.
\subsection{Loss Construction}

\noindent\textbf{Bipartite matching.}
After obtaining features from above-mentioned modules, Laneformer predicts a number of $N$ lane set according to its number of queries, where $N$ is set to be significantly larger than the maxinum number of lanes in the dataset. Therefore we need to pad the ground truth with non-lane to be a set of $N$ objects first, denoted as $\mathbf{G} = {\{g_n | g_n = (c_n, l_n)\}}_{n=1}^{N}$, where $c_n \in \{0, 1\}$, 0 represents non-lane and 1 represents lane. Given predicted outputs as $\mathbf{P} = {\{p_n | p_n = (\hat{p}_n, \hat{l}_n)\}}_{n=1}^{N}$, where $\hat{p}_n(c_n)$ stands for the probability score that $\hat{l}_n$ belongs to the specific category $c_n$.  We search for a permutation of $N$ elements matching index $\delta$ to minimize the pair-wise distance function $D$ between ground-truth lane $g_n$ and predicted lane $p_{\delta{(n)}}$:

\begin{equation}\label{best-matching}
\hat{\delta} = \mathop{\arg\min}_{\delta}{\sum_{n=1}^{N}D(g_n, p_{\delta{(n)}})}.
\end{equation}

The difference function $D$ is defined as following:
\begin{equation}
\label{difference function}
\fontsize{9.5pt}{\baselineskip}\selectfont{{D(g_n, p_{\delta{(n)}})\! = \!-\omega_{1}\hat{p}_{\delta{(n)}}(c_n) \!+\! \mathds{1}(c_{n} \!=\! 1)\mathcal{L}_{loc}(l_n, \hat{l}_{\delta{(n)}})}}
\end{equation}

Where $\mathds{1}(*)$ denotes an indicator function and $\omega_{1}$ stands for the coefficient of classification term.
$\mathcal{L}_{loc}$ is defined as follows:
\begin{equation}\label{distance}
\begin{aligned}
\mathcal{L}_{loc}(l_n, \hat{l}_{\delta{(n)}})\!=\! \omega_{2}&L_{1}(X_n, \hat{X}_{\delta{(n)}}) + \omega_{3}L_{1}(s_n, \hat{s}_{\delta{(n)}})\\ &+ \omega_{4}L_{1}(e_n, \hat{e}_{\delta{(n)}})
\end{aligned}
\end{equation}

\noindent where $L_1$ denotes the mean absolute error and $\omega_{2}$, $\omega_{3}$, $\omega_{4}$ indicate for the coefficients for point, start position and end position term respectively.
Bipartite matching is adopted to ensure one-to-one alignment between predictions and ground truths, making the Laneformer an end-to-end architecture by eliminating additional post-processing procedures.

\noindent\textbf{Total Loss}
The total loss of our model is calculated with matching index $\delta$ gained from bipartite matching, consisting of negative log-likelihood loss for classification prediction and $L1$-based location loss:
\begin{equation}
\fontsize{9.5pt}{\baselineskip}\selectfont{\mathcal{L}_{total} \!=\!  \sum_{n=1}^{N}-\omega_{1}log{\hat{p}_{\delta{(n)}}(c_n)} \!+\! \mathds{1}(c_n\!=\!1)\mathcal{L}_{loc}(l_n, \hat{l}_{\delta{(n)}})}
\end{equation}
Where $\mathcal{L}_{loc}$ is calculated the same with Eq.(\ref{distance}) and ($n$, $\delta{(n)}$) is the optimal pair indexes that minimize Eq.(\ref{best-matching}).
$\omega_{1}$, $\omega_{2}$, $\omega_{3}$, $\omega_{4}$ also adjust the effect of the loss terms and are set as same values with Eq.(\ref{difference function}) and Eq.(\ref{distance}).

%% file: file/4-experiment.tex
\section{Experiment}

\begin{table*}[ht!]
	\caption{\label{tab:Performance-on-CULane}Comparison of F1-measure(\%) and MACs(multiply–accumulate operations) on \textbf{CULane} testing set, where Laneformer* denotes for Laneformer without detection attention module. Our Laneformer achieves state-of-the-art performance.} %of the total test split.
	\resizebox{2.1\columnwidth}{!}{\tablestyle{8pt}{1.05}
	
	\begin{centering}
		{
			\begin{tabular}{c|ccccccccc|cc}
				%\hline 
				\shline 
				{\footnotesize{}Methods} & {\footnotesize{}Normal} & {\footnotesize{}Crowded} & {\footnotesize{}Dazzle} & {\footnotesize{}Shadow} & {\footnotesize{}No line} & {\footnotesize{}Arrow} & {\footnotesize{}Curve} & {\footnotesize{}Night} & {\footnotesize{}Cross}  & \textbf{\footnotesize{}Total} & {\footnotesize{}MACs (G)}\tabularnewline
				\shline
				{\footnotesize{}SCNN\cite{pan2018spatial}}& {\footnotesize{}90.60} & {\footnotesize{}69.70} & {\footnotesize{}58.50} & {\footnotesize{}66.90} & {\footnotesize{}43.40} & {\footnotesize{}84.10} & {\footnotesize{}64.40}  & {\footnotesize{}66.10}  & {\footnotesize{}1990} & {\footnotesize{}71.60} & \textbf{{\footnotesize{}/}} \tabularnewline
				%\hline 
				{\footnotesize{}ENet-SAD\cite{hou2019learning}} & {\footnotesize{}90.10} & {\footnotesize{}68.80} & {\footnotesize{}60.20} & {\footnotesize{}65.90} & {\footnotesize{}41.60} & {\footnotesize{}84.00} & {\footnotesize{}65.70}  & {\footnotesize{}66.00}  & {\footnotesize{}1998} & {\footnotesize{}70.80} & \textbf{{\footnotesize{}/}} \tabularnewline
				{\footnotesize{}PointLane\cite{chen2019pointlanenet}} & {\footnotesize{}88.00} & {\footnotesize{}68.10} & {\footnotesize{}61.50} & {\footnotesize{}63.30} & {\footnotesize{}44.00} & {\footnotesize{}80.90} & {\footnotesize{}65.20}  & {\footnotesize{}63.20}  & {\footnotesize{}1640} & {\footnotesize{}70.20} & \textbf{{\footnotesize{}/}}  \tabularnewline
				{\footnotesize{}ERFNet-HESA\cite{lee2021robust}}& \underline{\footnotesize{}92.00} & {\footnotesize{}73.10} & {\footnotesize{}63.80} & {\footnotesize{}75.00} & {\footnotesize{}45.00} & {\footnotesize{}88.20} & {\footnotesize{}67.90}  & {\footnotesize{}69.20}  & {\footnotesize{}2028}   & {\footnotesize{}74.20} & \textbf{{\footnotesize{}/}} \tabularnewline
				{\footnotesize{}CurveLane-S\cite{xu2020curvelane}} & {\footnotesize{}88.30} & {\footnotesize{}68.60} & {\footnotesize{}63.20} & {\footnotesize{}68.00} & {\footnotesize{}47.90} & {\footnotesize{}82.50} & {\footnotesize{}66.00}  & {\footnotesize{}66.20}  & {\footnotesize{}2817} & {\footnotesize{}71.40} & \textbf{{\footnotesize{}9.0}} \tabularnewline
				{\footnotesize{}CurveLane-M\cite{xu2020curvelane}} & {\footnotesize{}90.20} & {\footnotesize{}70.50} & {\footnotesize{}65.90} & {\footnotesize{}69.30} & {\footnotesize{}48.80} & {\footnotesize{}85.70} & {\footnotesize{}67.50}  & {\footnotesize{}68.20}  & {\footnotesize{}2359} & {\footnotesize{}73.50} & {\footnotesize{33.7}} \tabularnewline
				{\footnotesize{}CurveLane-L\cite{xu2020curvelane}} & {\footnotesize{}90.70} & {\footnotesize{}72.30} & {\footnotesize{}67.70} & {\footnotesize{}70.10} & \underline{\footnotesize{}49.40} & {\footnotesize{}85.80} & \textbf{\footnotesize{}68.40}  & {\footnotesize{}68.90}  & {\footnotesize{}1746} & {\footnotesize{}74.80} & {\footnotesize{86.5}}\tabularnewline
				{\footnotesize{}LaneATT(ResNet-18)\cite{tabelini2020keep}} & {\footnotesize{}91.17} & {\footnotesize{}72.71} & {\footnotesize{}65.82} & {\footnotesize{}68.03} & {\footnotesize{}49.13} & \underline{\footnotesize{}87.82} & {\footnotesize{}63.75}  & {\footnotesize{}68.58}  & {\footnotesize{}1020} & {\footnotesize{}75.13}  & \underline{\footnotesize{}9.3} \tabularnewline
				{\footnotesize{}LaneATT(ResNet-34)\cite{tabelini2020keep}} & \textbf{\footnotesize{}92.14} & {\footnotesize{}75.03} & {\footnotesize{}66.47} & \textbf{\footnotesize{}78.15} & {\footnotesize{}49.39} & \textbf{\footnotesize{}88.38} & {\footnotesize{}67.72}  & {\footnotesize{}70.72}  & {\footnotesize{}1330} & {\footnotesize{}76.68} & {\footnotesize{}18.0} \tabularnewline
				{\footnotesize{}LaneATT(ResNet-122)\cite{tabelini2020keep}} & {\footnotesize{}91.74} & \textbf{\footnotesize{}76.16} & \underline{\footnotesize{}69.47} & \underline{\footnotesize{}76.31} & \textbf{\footnotesize{}50.46} & {\footnotesize{}86.29} & {\footnotesize{}64.05}  & \underline{\footnotesize{}70.81}  & {\footnotesize{}1264} & \underline{\footnotesize{}77.02} & {\footnotesize{}70.5} \tabularnewline
				\hline 
				\textbf{\footnotesize{}Laneformer(ResNet-50)*} & {\footnotesize{}91.55} & {\footnotesize{}74.76} & {\footnotesize{}69.27} & {69.59} & {\footnotesize{}48.13} & {\footnotesize{}86.99} & \underline{\footnotesize{}68.15}  & {\footnotesize{}70.06}  & {\footnotesize{}1104} & {\footnotesize{}76.04} & {\footnotesize{}26.2} \tabularnewline
				\hline 
				\textbf{\footnotesize{}Laneformer(ResNet-18)} & {\footnotesize{}88.60} & {\footnotesize{}69.02} & {\footnotesize{}64.07} & {\footnotesize{}65.02} & {\footnotesize{}45.00} & {\footnotesize{}81.55} & {\footnotesize{}60.46}  & {\footnotesize{}64.76}  & \underline{\footnotesize{}25} & {\footnotesize{}71.71} & {\footnotesize{}13.8} \tabularnewline
				\textbf{\footnotesize{}Laneformer(ResNet-34)} & {\footnotesize{}90.74} & {\footnotesize{}72.31} & {\footnotesize{}69.12} & {\footnotesize{}71.57} & {\footnotesize{}47.37} & {\footnotesize{}85.07} & {\footnotesize{}65.90}  & {\footnotesize{}67.77}  & {\footnotesize{}26} & {\footnotesize{}74.70} & {\footnotesize{}23.0} \tabularnewline
				\textbf{\footnotesize{}Laneformer(ResNet-50)} & {\footnotesize{}91.77} & \underline{\footnotesize{}75.41} & {\textbf{\footnotesize{}70.17}} & {\footnotesize{}75.75} & {\footnotesize{}48.73} & {\footnotesize{}87.65} & {\footnotesize{}66.33}  & {\textbf{\footnotesize{}71.04}}  & {\textbf{\footnotesize{}19}} & {\textbf{\footnotesize{}77.06}} & {\footnotesize{}26.2}\tabularnewline
				\hline 
			\end{tabular}}{\scriptsize\par}
		\end{centering}}
	
\end{table*}
\subsection{Datasets and Evaluation Metrics}

We conduct experiments on the two most popular lane detection benchmarks.
\textbf{CULane} \cite{pan2018spatial} is a large-scale traffic lane detection dataset that is collected by in-vehicle cameras in Beijing, China.
It consists of 88,880 training images, 9675 validation
images, and 34,680 test images. The test split is further divided into
normal and 8 challenging categories. \textbf{TuSimple }\cite{TuSimple}
is an autonomous driving dataset which specifically
focuses on real highway scenarios, including 3626 images for training set and 2782 images for the test set.

\textbf{Evaluation Metrics}
We use F1 score(abbreviated as F1 in the following section) to measure the model performance on CULane: $F_{1}=\frac{2\times Precision\times Recall}{Precision+Recall}$, where $Precision=\frac{TP}{TP+FP}$ and $Recall=\frac{TP}{TP+FN}$. 
As for Tusimple, standard evaluation metrics including Accuracy, false positives(FP) and false negatives(FN) are adopted.

\subsection{Implementation Details}
The input resolution is set to $820 \times 295$ for CULane and $640 \times 360$ for TuSimple.
Data augmentations are applied on the raw image, consisting of horizontal flipping, a random choice from color-shifting operations(e.g., gaussian blur, linear contrast) and position-shifting operations(e.g., cropping, rotate).
Most of our experiments use ResNet50 as the backbone. We follow the setting of ~\cite{zhu2020deformable} and utilize a deformable transformer as the plain transformer. 
The bipartite matching and loss term coefficients $\omega_{1}$, $\omega_{2}$, $\omega_{3}$ and $\omega_{4}$ are set as 2, 10, 10, 10, respectively.  %and loss weight for class, point, start position and end position are 2, 5, 5 and 5.
Both the number of encoder and decoder layers is set to 1.
Moreover, we adopt 25 as the number of queries $N$ and 10 as the number of used detected bounding boxes $M$.
Eight V100s are used to train the model and the batch size is set to be 64. 
The learning rate is set to 1e-4 for the backbone and 1e-5 for the transformer.
We train 100 epochs on CULane and drop the learning rate by ten at 80 epoch.
On Tusimple, the total number of training iterations is set to 28k and the learning rate drops at 22k iteration.
During inference, the single scale test is adopted with the score threshold set to 0.8.
The trained detector used to obtain person-vehicle bounding boxes is based on common Faster-RCNN\cite{ren2015faster} architecture with ResNet-50 backbone and trained 12 epochs on BDD100K dataset\cite{yu2018bdd100k} with 70k images.

\subsection{Main Results}
\noindent\textbf{CULane.} Table \ref{tab:Performance-on-CULane} shows the Laneformer's performance on CULane test set. Our Laneformer achieves state-of-the-art results on F1 of the total test split. In addition, our Lanformer with only ResNet-50 backbone even surpasses the results of LaneATT~\cite{tabelini2020keep} with a larger ResNet-122 backbone.
Additionally, Laneformer outperforms all other lane detection models on some challenging splits such as ``Night", ``Dazzle light" and ``Cross".
Among the above three splits, \textbf{Laneformer significantly improves the performance of ``Cross" category, which achieves an extremely low FP(False Positive) number 19.}
We observe that our model gets a result of 1104 FP on ``Cross" category without detection attention, while with the detection attention, the performance makes a dramatic improvement as in Table \ref{tab:Performance-on-CULane}. The promotion may come from the perception of vehicle-person global context on the particular crossroad scenario benefiting from the proposed detection attention layer.

\noindent\textbf{TuSimple.} Experimental results on TuSimple benchmark are summarized in Table \ref{tab:Tusimple}.
We achieve 96.8\% accuracy, 5.6\% FP and 1.99\% FN with ResNet-50 backbone.
It is shown that Laneformer achieves comparable accuracy with the state-of-the-art Line-CNN and 0.6\% higher than another transformer-based method LSTR.
Even with the smaller backbones (ResNet-18, ResNet-34), Laneformer can achieve a competitive accuracy compared with the state-of-the-art methods.
Note that adding detection attention on Tusimple doesn't improve much due to the relatively simple highway driving scenes (e.g., few cars and straight lines).
	
\begin{table}[h!]
	\caption{\label{tab:Tusimple}Comparison of different algorithms on the \textbf{Tusimple} testing benchmark, where Laneformer* denotes for Laneformer without detection attention module.}
	\resizebox{1.0\columnwidth}{!}{\tablestyle{8pt}{1.05}
	\begin{centering}
		{
			\begin{tabular}{cccc}
			    \shline 
				{\footnotesize{}Method}  & {\footnotesize{}Acc(\%)} & {\footnotesize{}FP(\%)} & {\footnotesize{}FN(\%)} \tabularnewline
				\shline 
				{\footnotesize{}SCNN \cite{pan2018spatial}}  & {\footnotesize{}96.53} & {\footnotesize{}6.17} & \textbf{\footnotesize{}1.80}\tabularnewline
				{\footnotesize{}LSTR \cite{liu2021end}}  & {\footnotesize{}96.18} & \textbf{\footnotesize{}2.91} & {\footnotesize{}3.38}\tabularnewline
				{\footnotesize{}Enet-SAD \cite{hou2019learning}} & {\footnotesize{}96.64} & {\footnotesize{}6.02} & {\footnotesize{}2.05}\tabularnewline
				{\footnotesize{}Line-CNN \cite{li2019line}} & \textbf{\footnotesize{}96.87} & {\footnotesize{}4.41} & {\footnotesize{}3.36}\tabularnewline
				{\footnotesize{}PolyLaneNet \cite{tabelini2020polylanenet}}  & {\footnotesize{}93.36} & {\footnotesize{}9.42} & {\footnotesize{}9.33}\tabularnewline
				{\footnotesize{}PointLaneNet \cite{chen2019pointlanenet}} & {\footnotesize{}96.34} & {\footnotesize{}4.67} & {\footnotesize{}5.18}\tabularnewline
				{\footnotesize{}LaneATT (ResNet-18) \cite{tabelini2020keep}} & {\footnotesize{}95.57} & {\footnotesize{}3.56} & {\footnotesize{}3.01}\tabularnewline
				{\footnotesize{}LaneATT (ResNet-34) \cite{tabelini2020keep}} & {\footnotesize{}95.63} & {\footnotesize{}3.53} & {\footnotesize{}2.92}\tabularnewline
				{\footnotesize{}LaneATT (ResNet-122) \cite{tabelini2020keep}} & {\footnotesize{}96.10} & {\footnotesize{}5.64} & {\footnotesize{}2.17}\tabularnewline
				\hline 
				\textbf{\footnotesize{}Laneformer(ResNet-50)*}  & {\footnotesize{}96.72} & \underline{\footnotesize{}3.46} & {\footnotesize{}2.52}\tabularnewline
				\hline 
				\textbf{\footnotesize{}Laneformer(ResNet-18)} & {\footnotesize{}96.54} & {\footnotesize{}4.35} & {\footnotesize{}2.36} \tabularnewline
				\textbf{\footnotesize{}Laneformer(ResNet-34)} & {\footnotesize{}96.56} & {\footnotesize{}5.39} & {\footnotesize{}3.37} \tabularnewline
				\textbf{\footnotesize{}Laneformer(ResNet-50)}  & \textbf{\footnotesize{}96.80} & {\footnotesize{}5.60} & \underline{\footnotesize{}1.99}\tabularnewline
				\hline 
			\end{tabular}{\footnotesize\par}}
		\end{centering}}
	
\end{table}	

\noindent\textbf{Latency Comparison.} 
In inference, Laneformer with ResNet-50 backbone achieves 53 and 48 FPS on one V100 for CULane and Tusimple benchmark.
For latency comparison of different components in Laneformer, we conduct experiments on CULane testing split and illustrate the result in Table \ref{tab:ablation_components1}.
After adding row-column attention and detection attention, there is only a 4.9\%, 8.1\% increment on inference FPS due to the efficient matrix multiplication. 

\begin{figure*}[ht!]
    \label{fig:attention map3}
	\centering
		\subfigure[Visualization of detection attention.]{                 \includegraphics[scale=0.39]{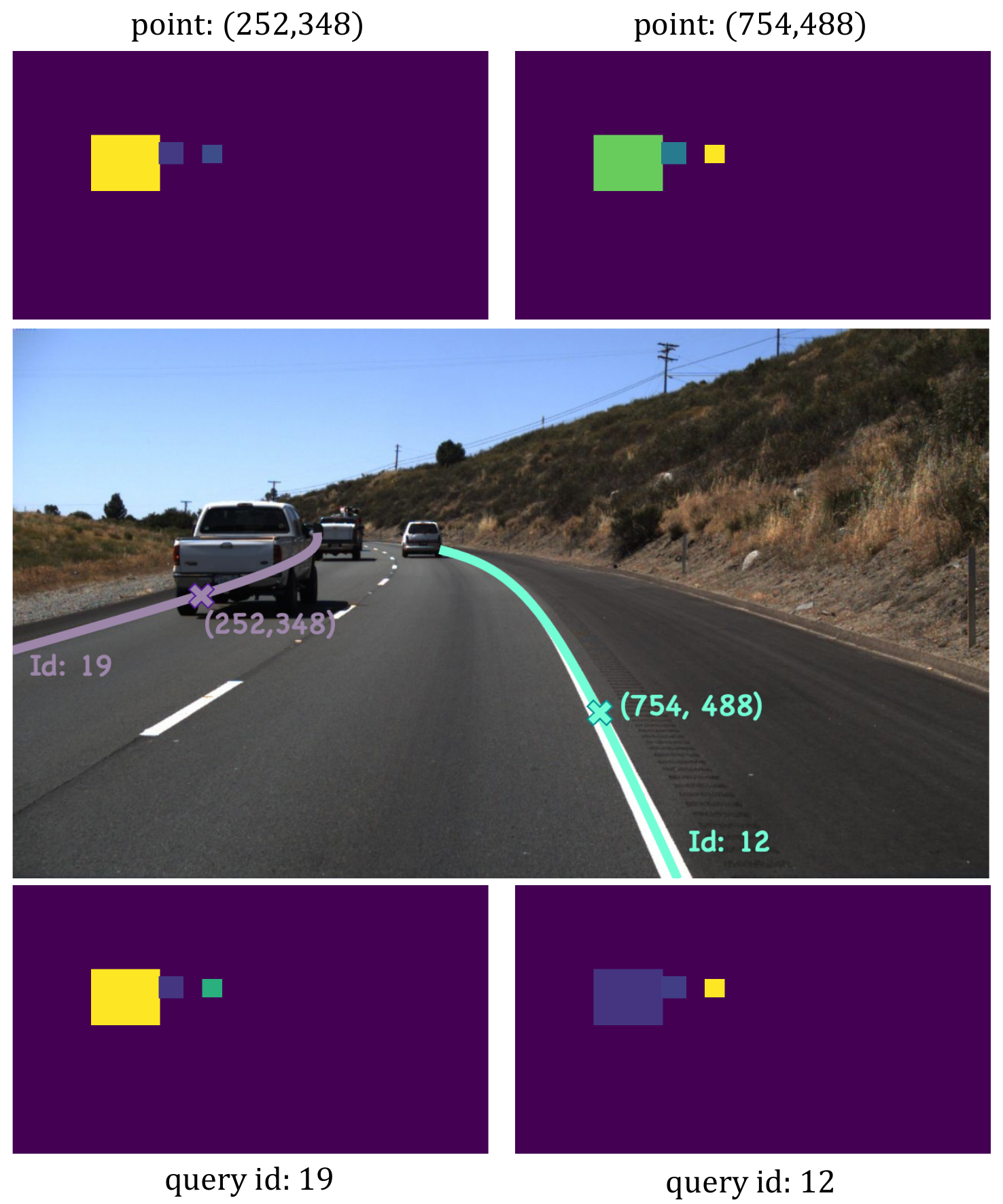}
		    \label{fig:visualization-detection-attention}}
		\hspace{0.9in}
		\subfigure[Visualization of row and column attention.]{           \includegraphics[scale=0.39]{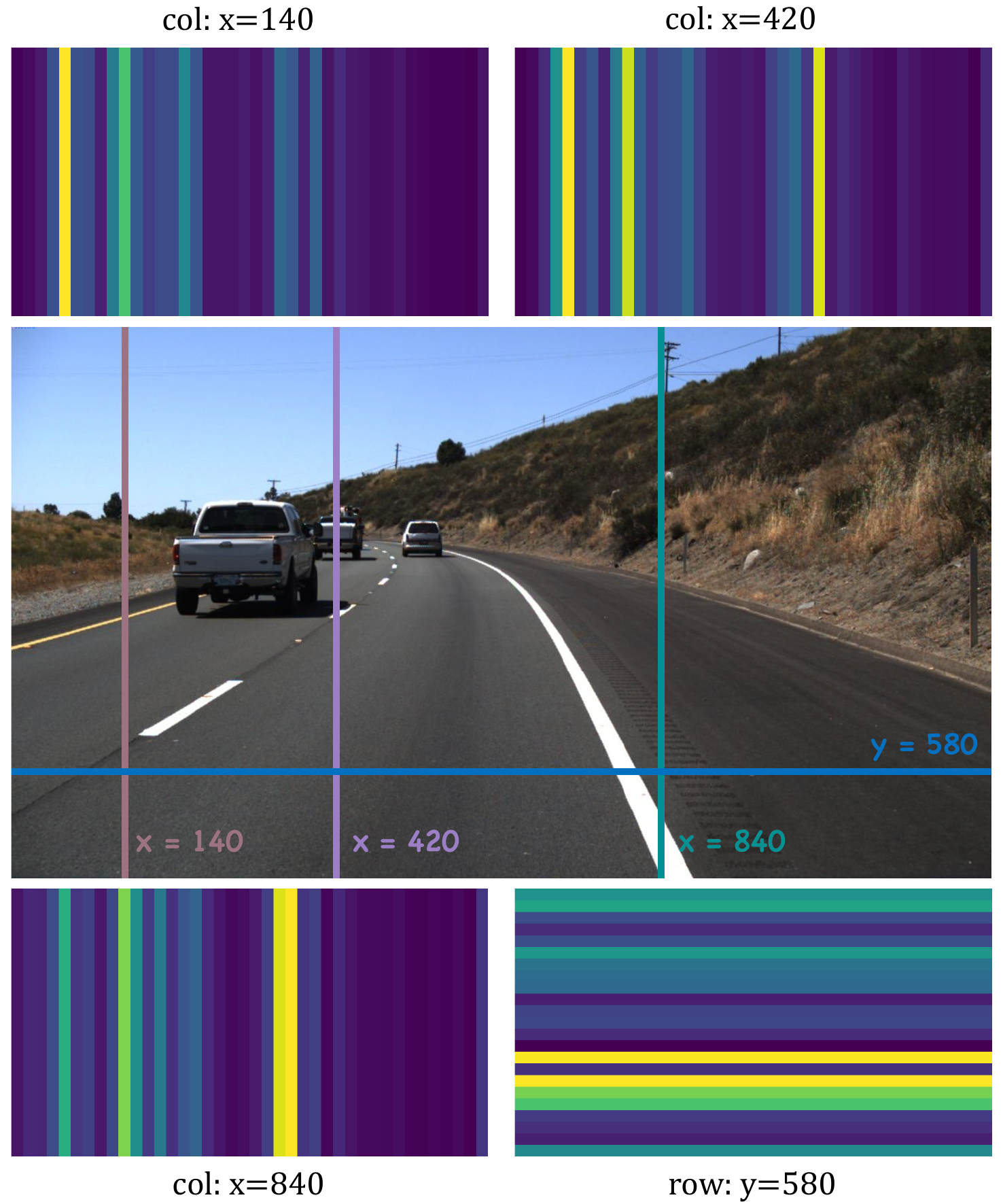} 
		  \label{fig:visualization-row-column-attention}} 
	\caption{Visualization of proposed attentions on Tusimple dataset, where brighter color denotes more significant attention value. (a) shows the detection attention map for different points and queries in the transformer. We can observe either the pixel or the query focus on the detection bounding boxes near the corresponding lane. (b) shows that the row attention mainly considers the nearby rows, while the column attention focuses on the nearby representative column of each lane.}
\end{figure*}

\noindent\textbf{Visualization.}
We visualize several attention maps in the transformer to find out the area that detection attentions and row-column attentions focus on, where brighter color denotes for more significant attention value.
Shown in Figure \ref{fig:visualization-detection-attention}, either the point or query pays more attention to the detected instances besides lanes it responsible for, especially when those instances are in occlusion with part of the lane. 
Moreover, observation can be made in Figure \ref{fig:visualization-row-column-attention} that row attention mainly considers nearby rows, while the column attention focuses on the nearby representative column of each lane.
These results demonstrate our assumption that the implicit relationship of traffic scenes can be obtained from proposed detection attention and row-column attention.
%The visualization of original attention in transformer can be referred to supplementary materials.

\subsection{Ablation Study}
% In this section, we conduct ablation studies to provide more insights into different components and settings for Laneformer. 
% Ablation studies are conducted on the CULane dataset as it is more diverse and challenging. 

\noindent\textbf{Different components.} As we can see in Table \ref{tab:ablation_components1}, without detection attention and row-column attention, our baseline (plain transformer) obtains 75.45\% on F1. 
The adding of row-column attention improves the performance to 76.04\%.
What's more, simply introducing detected objects information can improve the performance of our model, but a little more extra information such as scores or categories will make it better. 
We can observe that making full use of detected objects with their scores and categories raises the F1 to 77.06\%, which is the state-of-the-art results on CULane.

\begin{table}[htb]
	\caption{\label{tab:ablation_components1} Ablation study results of different components of Laneformer on \textbf{CULane}. }
	\resizebox{1.0\columnwidth}{!}{\tablestyle{4pt}{1.05}
	\begin{centering}
		{
			\begin{tabular}{l|ccc|cc}
				\shline 
				{\footnotesize{}Model} & {\footnotesize{}F1(\%)} & {\footnotesize{}Precision(\%)} & {\footnotesize{}Recall(\%)} & {\footnotesize{}FPS} &{\footnotesize{}Params. (M)}\tabularnewline
				\shline 
				{\footnotesize{}Baseline(ResNet-50)} & {\footnotesize{}75.45} & {\footnotesize{}81.65} & {\footnotesize{}70.11} &  {\footnotesize{}61} &  {\footnotesize{}31.02}\tabularnewline
				\hline 
				{\footnotesize{}+ row-column attention} & {\footnotesize{}76.04} & {\footnotesize{}82.92} & {\footnotesize{}70.22}&  {\footnotesize{}58}&  {\footnotesize{}43.02}\tabularnewline
				{\footnotesize{}+ bounding box} & {\footnotesize{}76.08} & {\footnotesize{}85.30} & {\footnotesize{}68.66}&  {\footnotesize{}57} &  {\footnotesize{}45.38}\tabularnewline
				{\footnotesize{}+ score} & {\footnotesize{}76.25} & {\footnotesize{}83.56} & {\footnotesize{}70.12}&  {\footnotesize{}54} &  {\footnotesize{}45.38}\tabularnewline
				%{\footnotesize{}+ category} & {\footnotesize{}76.35} & {\footnotesize{}85.40} & {\footnotesize{}69.03}\tabularnewline
				{\footnotesize{}+ category} & {\footnotesize{}77.06} & {\footnotesize{}84.05} & {\footnotesize{}71.14} &  {\footnotesize{}53} &  {\footnotesize{}45.38}\tabularnewline
			
				\hline 
			\end{tabular}{\footnotesize\par}}
		\par\end{centering}}
\end{table}

\noindent\textbf{Score threshold of the bounding box.} To find out the influences of detected objects with different confidence, we conduct a series of experiments with different score thresholds on filtering detected bounding boxes.
Table \ref{tab:ablation_components2} shows that using detection outputs from different score thresholds has a slight impact on our results. 
Especially, results between threshold 0.6 to 0.8 are robust and the differences can be nearly neglect. 
When the threshold is lower than 0.6, there may be some noise in detected bounding boxes and therefore our performance gets slighted hurt.
On the other hand, if the threshold is too high such as 0.9, only a few bounding boxes will be chosen so that a large number of padded boxes will interfere with the model's learning, which leads to a lower F1.
Besides, we experiment on random bounding boxes to prove that Laneformer indeed makes use of the information of detected objects. 

\begin{table}[tb]
	\caption{\label{tab:ablation_components2} Quantitative evaluation of different detection bounding box input settings on \textbf{CULane} testing split. }
	\resizebox{1.0\columnwidth}{!}{\tablestyle{4pt}{1.05}
	\begin{centering}
		{
			\begin{tabular}{l|l|ccc}
				\shline 
				{\footnotesize{}} & {\footnotesize{}} & {\footnotesize{}F1(\%)} & {\footnotesize{}Precision(\%)} & {\footnotesize{}Recall(\%)} \tabularnewline
				\shline 
                \multirow{6}*{\footnotesize{}Score threshold} &{\footnotesize{}0.4} & {\footnotesize{}76.34} & {\footnotesize{}85.41} & {\footnotesize{}69.01}\\
                ~ &{\footnotesize{}0.5} & {\footnotesize{}76.14} & {\footnotesize{}84.52} & {\footnotesize{}69.27}\\
                ~ &{\textbf{\footnotesize{}0.6}} & {\textbf{\footnotesize{}77.06}} & {{\footnotesize{}84.05}} & {\textbf{\footnotesize{}71.14}}\\
                ~ &{\footnotesize{}0.7} & {\footnotesize{}76.71} & {\footnotesize{}84.53} & {\footnotesize{}70.21}\\
                ~ &{\footnotesize{}0.8} & {\footnotesize{}76.85} & {\footnotesize{}84.51} & {\footnotesize{}70.46}\\
                ~ &{\footnotesize{}0.9} & {\footnotesize{}76.37} & \textbf{\footnotesize{}85.54} & {\footnotesize{}68.98}\\
                \cline{2-5}
                ~ &{\footnotesize{}random} & {\footnotesize{}75.99} & {\footnotesize{}84.50} & {\footnotesize{}69.03}\\
	          %  \cline{2-4}
				\hline 
				\multirow{3}*{\footnotesize{}Number of Bbox} &{\footnotesize{}5} & {\footnotesize{}76.90} & {\footnotesize{}84.30} & {\footnotesize{}70.69}\\
				~ &{\textbf{\footnotesize{}10}} & {\textbf{\footnotesize{}77.06}} & {{\footnotesize{}84.05}} & {\textbf{\footnotesize{}71.14}}\\
                ~ &{\footnotesize{}20} & {\footnotesize{}76.51} & \textbf{\footnotesize{}84.34} & {\footnotesize{}70.01}\\
                \hline 
				\multirow{4}*{\footnotesize{}Different categories} &{\footnotesize{}none} & {\footnotesize{}76.04} & {\footnotesize{}82.92} & {\footnotesize{}70.22}\\
				~ &{\footnotesize{}person} & {\footnotesize{}76.40} & \textbf{\footnotesize{}85.17} & {\footnotesize{}69.27}\\
				~ &{\footnotesize{}vehicle} & {\footnotesize{}76.79} & {\footnotesize{}84.44} & {\footnotesize{}70.41}\\
                ~ &{\textbf{\footnotesize{}all}} & {\textbf{\footnotesize{}77.06}} & {{\footnotesize{}84.05}} & {\textbf{\footnotesize{}71.14}}\\
				\hline 
			\end{tabular}{\footnotesize\par}}
		\par\end{centering}}
\end{table}

\noindent\textbf{Number of bounding box.} Apart from thresholds, we further explore the impact of using a different number of detected objects in Laneformer. The results in Table \ref{tab:ablation_components2} show that too many detected objects lead to a performance drop, and our model reaches the best result under the setting of 10 bounding boxes.
We speculate that the best setting of 10 bounding boxes is due to the average number of detected objects in each image under the bounding box threshold of 0.6, which is 9.84. 
So if we set a number much larger than 9.84, for example, 20 in our experiment, then too many useless padded boxes will be used, which may hurt the model's performance.
On the other hand, if we use too few bounding boxes, the information of detected objects is not entirely utilized to reach the best performance.

\noindent\textbf{Different categories.} Vehicles and persons have different relations with lanes. 
To be specific, vehicles are on the road, beside the lanes, while persons usually far from lanes. 
So we also explore the impact of extra information with different categories of bounding boxes.
Results in Table \ref{tab:ablation_components2} show that both the adding of vehicles and persons can improve the model performance, and the result with vehicles is better than the one with persons.
We suppose that vehicles share a more close relation with lanes in the perspective of locations.
Experiments show that the model with all of the two categories reaches the best results.

%% file: file/5-conclusion.tex
\section{Conclusions}
In this work, we propose Laneformer, a conceptually simple yet powerful transformer-based architecture tailored for lane detection.
Equipped with row and column self-attention module and semantic contexts provided by additional detected object instances, Laneformer achieves the state-of-the-art performance, in terms of 77.1\% F1 score on CULane and superior 96.8\% Accuracy on TuSimple benchmark.
Besides, visualization of the learned attention map in transformer demonstrates that our Laneformer can incorporate relations of long-range lane points and global contexts induced by surrounding objects.

%% file: file/6-appendix.tex
\section{Appendix}

We conduct more ablation studies to confirm the effectiveness of Laneformer. In the following section, we will show the results of these experiments based on ResNet-50.

\subsection{Layers of Encoder and Decoder}

It can be seen in Table.\ref{tab:enc_dec}, the different number of encoder layers and decoder layers have a slight difference in performances, which proves that our Laneformer is robust. We can observe that the increment of encoder layers and decoder layers can improve the performance on Recall metric.

\begin{table}[h]
	\caption{\label{tab:enc_dec}Experiments on different numbers of encoder and decoder layers on \textbf{CULane}.}
	\resizebox{1.0\columnwidth}{!}{\tablestyle{8pt}{1.05}
	\begin{centering}
		\setlength{\tabcolsep}{1mm}
		{
			\begin{tabular}{cc|ccc}
			    \shline 
				{\footnotesize{}Encoder} & {\footnotesize{}Decoder}& {\footnotesize{}F1(\%)}  & {\footnotesize{}Precision(\%)} & {\footnotesize{}Recall(\%)}  \tabularnewline
				\shline 
				{\footnotesize{}1} & {\footnotesize{}1} & \textbf{\footnotesize{}77.06}  & {\footnotesize{}84.05} & {\footnotesize{}71.14}\tabularnewline
				{\footnotesize{}2} & {\footnotesize{}2} & {\footnotesize{}76.48}  & {\footnotesize{}82.49} & {\footnotesize{}71.28} \tabularnewline
				{\footnotesize{}3} & {\footnotesize{}3} & {\footnotesize{}76.91}  & {\footnotesize{}82.59} & {\footnotesize{}71.97} \tabularnewline
				{\footnotesize{}1} & {\footnotesize{}2} & {\footnotesize{}76.90} & {\footnotesize{}82.92} & {\footnotesize{}71.70} \tabularnewline
				{\footnotesize{}1}  & {\footnotesize{}3} & {\footnotesize{}76.71} & {\footnotesize{}82.95} & {\footnotesize{}71.35} \tabularnewline
				{\footnotesize{}2} & {\footnotesize{}1} & {\footnotesize{}76.07} & {\footnotesize{}80.13} & {\footnotesize{}72.39} \tabularnewline
				{\footnotesize{}3} & {\footnotesize{}1} & {\footnotesize{}76.13} & {\footnotesize{}81.87} & {\footnotesize{}71.14} \tabularnewline
				\hline 
			\end{tabular}{\footnotesize\par}}
	\end{centering}}
\end{table}

\subsection{Threshold and Bounding Box Number}
As we have mentioned in our paper, the number of bounding boxes and the score threshold of bounding boxes have a relationship that they will influence the choice of each other. We do another two experiments on threshold 0.4 and 0.9 to verify this influence. First, we count the average objects number in each image under threshold 0.4 and 0.9, which is 11 objects and 7 objects, respectively. Then, we conduct experiments in two settings. Concretely, we use 11 bounding boxes for 0.4 score threshold and 7 bounding boxes for 0.9  score threshold. As we can see in Table.\ref{tab:threshold_number}, using the proper number of bounding boxes under corresponding threshold can clearly improve the performances.

\begin{table}[h]
	\caption{\label{tab:threshold_number} Experiments on different score threshold and bounding box number on \textbf{CULane}. }
	\resizebox{1.0\columnwidth}{!}{\tablestyle{8pt}{1.05}
	\begin{centering}
		{
			\begin{tabular}{l|l|ccc}
				\shline 
				{\footnotesize{}Threshold} & {\footnotesize{}Number} & {\footnotesize{}F1(\%)} & {\footnotesize{}Precision(\%)} & {\footnotesize{}Recall(\%)} \tabularnewline
				\shline 

                \multirow{2}*{\footnotesize{}0.4} &{\footnotesize{}10} & {\footnotesize{}76.34} & {\footnotesize{}85.41} & {\footnotesize{}69.01}\\
                ~ &{\footnotesize{}11} & \textbf{\footnotesize{}76.77} & {\footnotesize{}84.88} & {\footnotesize{}70.08}\\
	          %  \cline{2-4}
				\hline 
				\multirow{2}*{\footnotesize{}0.9} &{\footnotesize{}10} & {\footnotesize{}76.37} & {\footnotesize{}85.54} & {\footnotesize{}68.98}\\
				~ &{\footnotesize{}7} & \textbf{\footnotesize{}76.46} & {{\footnotesize{}84.61}} & {\footnotesize{}69.74}\\
				\hline 
			\end{tabular}{\footnotesize\par}}
		\par\end{centering}}
\end{table}
\subsection{Different Detectors}
To find out the influence of person-vehicle detectors with different accuracy, we also conduct experiments on different Faster-RCNNs trained on 10k, 30k, 70k images chosen from BDD100K dataset. Random bounding box setting is added for comparison.
Results in Table.\ref{tab:more_ablation} demonstrate that our Laneformer can work well with detectors of different accuracy.
Moreover, with the improvement of detector's accuracy, the performance of Laneformer also improves.

\begin{table}[ht]
	\caption{\label{tab:more_ablation} Experiments on different detectors, ROI sizes and different layers on \textbf{CULane}. }
	\resizebox{1.0\columnwidth}{!}{\tablestyle{4pt}{1.05}
	\begin{centering}
		{
			\begin{tabular}{l|l|ccc}
				\shline 
				{\footnotesize{}} & {\footnotesize{}} & {\footnotesize{}F1(\%)} & {\footnotesize{}Precision(\%)} & {\footnotesize{}Recall(\%)} \tabularnewline
				\shline 
				%{\footnotesize{}Laneformer} & &{\footnotesize{}77.06} & {\footnotesize{}/\dxj{add}} & {\footnotesize{}/\dxj{add}}\tabularnewline
				%\hline 
                \multirow{2}*{\footnotesize{}Different Detectors} &{\footnotesize{}random} & {\footnotesize{}75.99} & {\footnotesize{}84.50} & {\footnotesize{}69.03}\\
                ~ &{\footnotesize{}10k} & {\footnotesize{}76.71} & {\footnotesize{}85.36} & {\footnotesize{}69.65}\\
                ~ &{\footnotesize{}30k} & {\footnotesize{}76.96} & {\footnotesize{}85.10} & {\footnotesize{}70.25}\\
                ~ &{\footnotesize{}70k} & \textbf{\footnotesize{}77.06} & {\footnotesize{}84.05} & {\footnotesize{}71.14}\\
				\hline 
				\multirow{3}*{\footnotesize{}ROI size} &{\footnotesize{}$3\times 3$} & \textbf{\footnotesize{}77.06} & {\footnotesize{}84.05} & {\footnotesize{}71.14}\\
				~ &{\footnotesize{}$5\times 5$} & {\footnotesize{}76.06} & {\footnotesize{}83.80} & {\footnotesize{}69.62}\\
				~ &{\footnotesize{}$7\times 7$} & {\footnotesize{}76.05} & {{\footnotesize{}80.72}} & {\footnotesize{}71.88}\\
                \hline 
				\multirow{3}*{\footnotesize{}Different Layers} &{\footnotesize{}C3} & {\footnotesize{}76.47} & {\footnotesize{}85.76} & {\footnotesize{}69.00}\\
				~ &{\footnotesize{}C4} & \textbf{\footnotesize{}77.06} & {\footnotesize{}84.05} & {\footnotesize{}71.14}\\
				~ &{\footnotesize{}C5} & {\footnotesize{}75.94} & {\footnotesize{}80.75} & {\footnotesize{}71.67}\\
				\hline 
			\end{tabular}{\footnotesize\par}}
		\par\end{centering}}
\end{table}
\subsection{ROI Sizes}

We can observe from Table.\ref{tab:more_ablation} that the bigger the extracted ROI features, the lower of F1 score. We speculate that we extract ROI features from C4 layer of backbone features, which is of size $52 \times 19$ that is very small, so ROI features of size $3 \times 3$ is enough. If the size goes too big, the extracted feature needs to be interpolated to the specified size, which may introduce fuzzy information.

\subsection{ROI Features from Different Layers}
Our Laneformer extracts multi-stage features from the backbone, so we have choices to extract ROI features from backbone feature of different layers. We try for C3, C4 and C5 layers. Table.\ref{tab:more_ablation} shows that we get our best performance on ROI features extracted from C4 layer. While features extracted from C5 layer show a significant performance drop due to the feature size is too small that it loss too much information.
% \section{More Visualization Results}

\begin{figure*}[t]
    \label{fig:attention map2}
	\centering
	    \subfigure[\small{}Lane detection results on \textbf{TuSimple} dataset.]{           \includegraphics[scale=0.6]{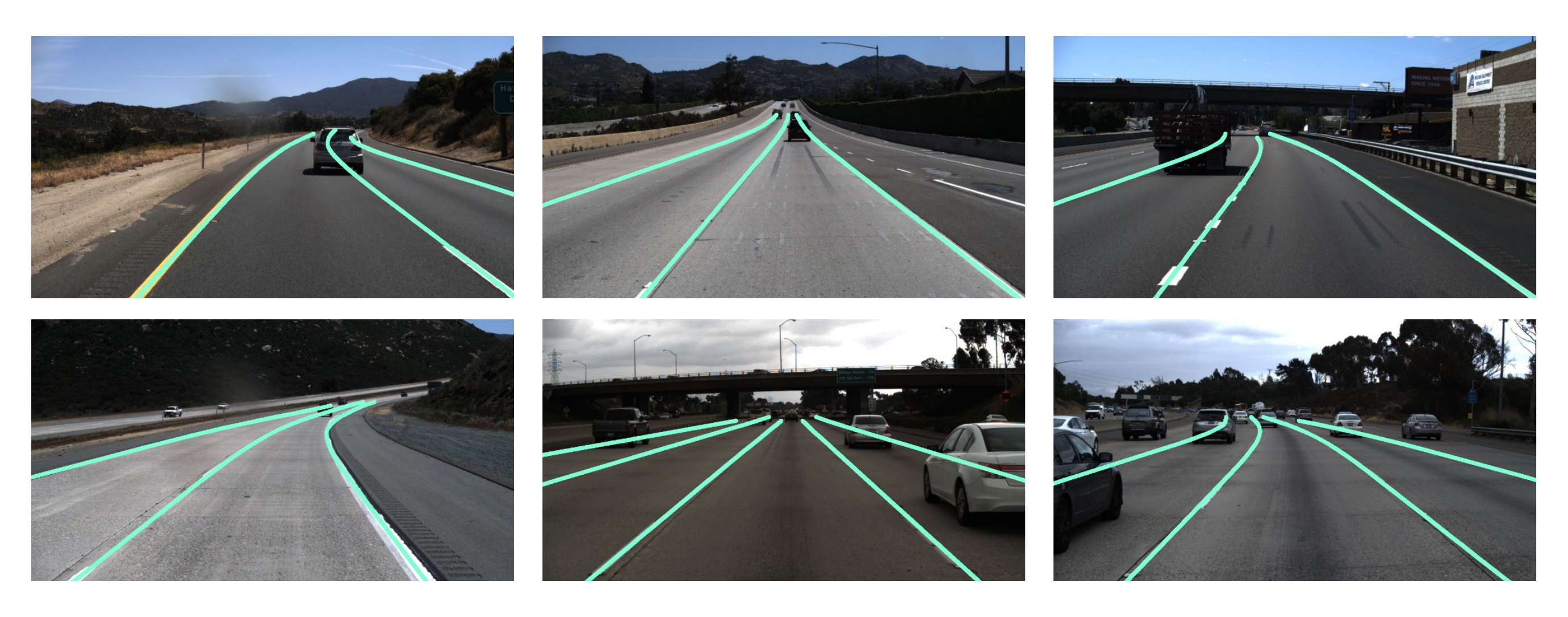} 
		\label{fig:Tusimple_visual}}
		\subfigure[\small{}Lane detection results on \textbf{CULane} dataset.]{                 \includegraphics[scale=0.6]{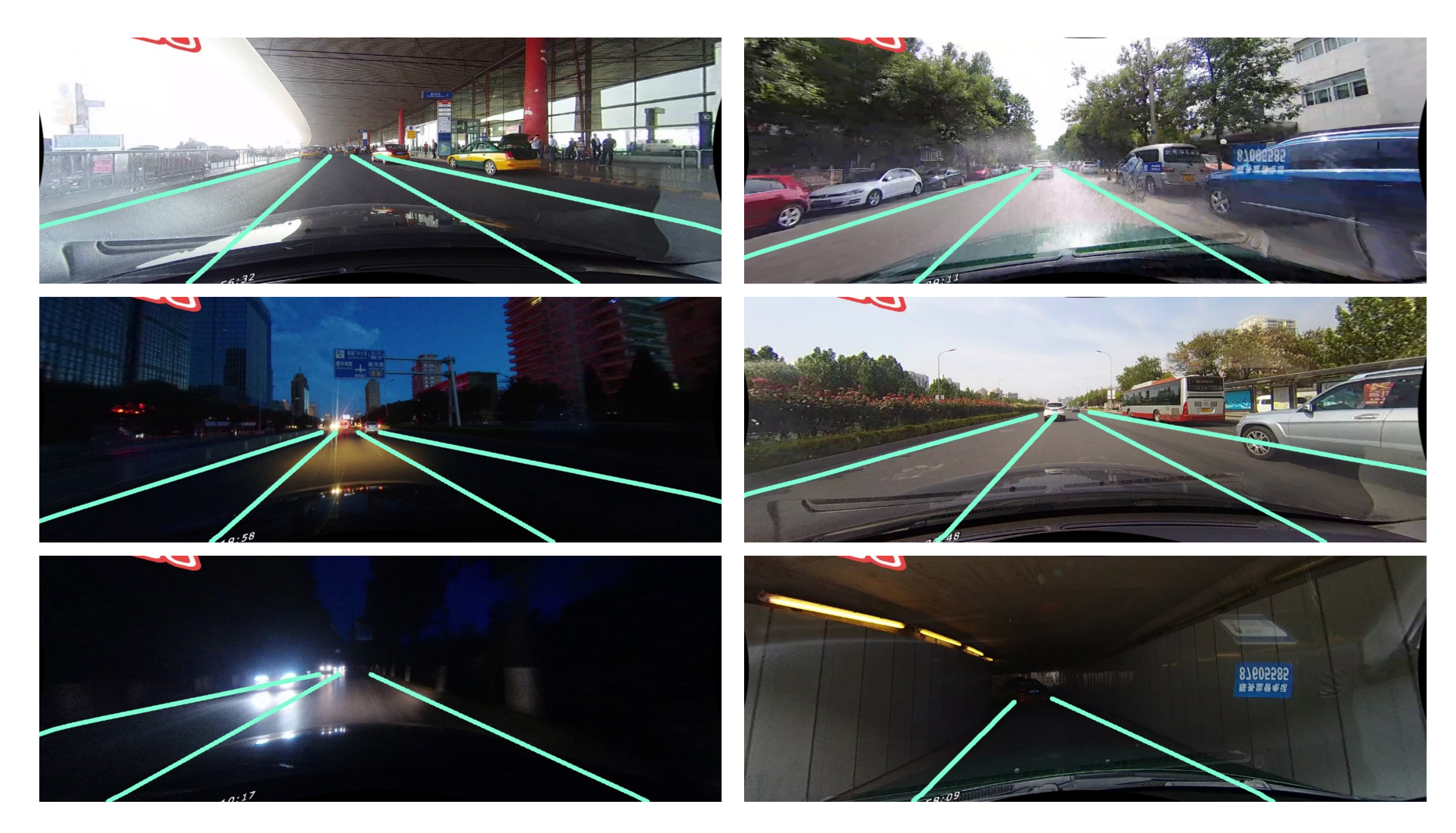} 
		\label{fig:CULane_visual}}
	\caption{Visualization of Laneformer results on \textbf{TuSimple} and \textbf{CULane} dataset.}%\xd{add details about the difference with previous attention}
	
\end{figure*}